\documentclass[runningheads]{llncs}

\usepackage{graphicx}
\usepackage{comment}
\usepackage{amsmath,amssymb}
\usepackage{color}
\usepackage{url}

\usepackage{mathtools}
\usepackage{booktabs}
\usepackage{multirow}
\usepackage{subcaption}

\usepackage{pgfplots}
\usepgfplotslibrary{groupplots}

\usepackage[section]{placeins}

\newcommand{\eg}{e.g.\ }
\newcommand{\ie}{i.e.\ }
\newcommand{\etal}{et al.\ }

\usepackage[pagebackref=true,breaklinks=true,letterpaper=true,colorlinks,bookmarks=false]{hyperref}

\DeclarePairedDelimiterX{\abs}[1]{\lvert}{\rvert}{#1}%
\DeclarePairedDelimiter\norm{\lVert}{\rVert}%

\pgfplotsset{compat=1.18}

\begin{document}

\title{A Study of Forward-Forward Algorithm for Self-Supervised Learning}
\author{Jonas Brenig\orcidID{0009-0004-3770-9643} \and
Radu Timofte\orcidID{0000-0002-1478-0402}}

\authorrunning{J. Brenig, R. Timofte}

\institute{Computer Vision Lab, CAIDAS \& IFI, University of W\"urzburg, Germany
\email{\{jonas.brenig,radu.timofte\}@uni-wuerzburg.de}}

\maketitle              

\begin{abstract}
    Self-supervised representation learning has seen remarkable progress in the last few years, with some of the recent methods being able to learn useful image representations without labels. 
    These methods are trained using backpropagation, the \textit{de facto} standard.
    Recently, Geoffrey Hinton proposed the forward-forward algorithm as an alternative training method. It utilizes two forward passes and a separate loss function for each layer to train the network without backpropagation. 
    
    In this study, for the first time, we study the performance of forward-forward vs. backpropagation for self-supervised representation learning and provide insights into the learned representation spaces.
    Our benchmark employs four standard datasets, namely MNIST, F-MNIST, SVHN and CIFAR-10, and three commonly used self-supervised representation learning techniques, namely rotation, flip and jigsaw.
    
    Our main finding is that while the forward-forward algorithm performs comparably to backpropagation during (self-)supervised training, the transfer performance is significantly lagging behind in all the studied settings. This may be caused by a combination of factors, including having a loss function for each layer and the way the supervised training is realized in the forward-forward paradigm. In comparison to backpropagation, the forward-forward algorithm focuses more on the boundaries and drops part of the information unnecessary for making decisions which harms the representation learning goal. Further investigation and research are necessary to stabilize the forward-forward strategy for self-supervised learning, to work beyond the datasets and configurations demonstrated by Geoffrey Hinton. 
\end{abstract}

\section{Introduction}

While there are a lot of ways to do machine learning, for computer vision tasks neural networks have established themselves as one of the strongest methods, able to surpass most handcrafted methods. As the name suggests, the use of neural networks allows for some analogies to the biological brain. While there are quite a number of obvious differences, the overall structure seems somewhat comparable. 
One of the most interesting areas to compare between the two might be the way they learn.

At this time, almost every artificial neural network is trained using backpropagation algorithm (BP)~\cite{rumelhart1986learning}.
However, while backpropagation works well for artificial neural networks, research indicates that the biological brain does not use such a mechanism for learning~\cite{lillicrap2020backpropagation}. There seems to be no evidence that supports the idea of such a mechanism for biological brains. Naturally, the question arises whether it is possible to train neural networks without backpropagation.

The forward-forward algorithm (FF) is an alternative training mechanism that replaces backpropagation (BP) by several forward passes and was proposed recently by Geoffrey Hinton~\cite{hinton2022forward}. 
Inspired by earlier works like Boltzmann machines~\cite{hinton1986learning} and Noise Contrastive Estimation~\cite{gutmann2010noise}, it replaces backpropagation by forwarding through the network once with \emph{positive} data and once with \emph{negative} data, learning each layer separately to distinguish between both kinds of data. Learning happens by comparing layer activations, targeting lower activity for \emph{negative} data, and higher activity for \emph{positive} data.
Hinton experimented using this approach for a classification task, using supervised, as well as unsupervised learning approaches. 

One area of computer vision that has seen a lot of success in the last few years is self-supervised representation learning (SSL)~\cite{swav,dino,simclr,simsiam,byol,mocov1,jaiswal2020survey}. These approaches can learn from large unlabeled datasets, only requiring some fine-tuning of the network on the final task. As labeled datasets can be quite expensive to create, the idea of learning image representations without labels has been a popular area of research~\cite{goyal2021self,welinder2010online}.

Some of the recent works in this area use forms of a contrastive loss (\eg SimCLR~\cite{simclr}, BarlowTwins~\cite{barlowtwins} or MoCo~\cite{mocov2,mocov3,mocov1}), which train the network to distinguish different groups of samples. There has also been some success to eliminate the explicit contrastive loss (\eg BYOL~\cite{byol} or SimSiam~\cite{simsiam}).
While these approaches operate on different (augmented) views of the same image, there are also methods that operate on a single view of an image to solve tasks like predicting rotation~\cite{rotation} or solving puzzles~\cite{chen2021jigsaw,doersch2015unsupervised,noroozi2016unsupervised}.

In this paper, we study, for the first time, the performance of the forward-forward algorithm when applied to self-supervised representation learning to determine possible advantages and drawbacks this new method of training has in this setting. 
While Hinton proposed a specific way to train a network in an unsupervised manner, we investigate how some well-known approaches to self-supervised learning could be applied.
For this purpose, our benchmark compares three self-supervised representation learning techniques, namely rotation, flip and jigsaw, on four different datasets, namely MNIST, F-MNIST, SVHN and CIFAR-10. 
We aim to provide insights into differences of how the forward-forward algorithm learns, compared to backpropagation. 
Specifically, we compare the embedding space of networks trained using both methods.

The paper is structured as follows. 
First, we review Hinton's~\cite{hinton2022forward} forward-forward algorithm in Section~\ref{sec:forward-forward}, then the self-supervised representation learning in Section~\ref{sec:SSL}. In Section~\ref{sec:experimental_setup} we describe the experimental setup, the datasets and the techniques employed, while in Section~\ref{sec:results} the experimental results are reported and discussed. We draw the conclusions in Section~\ref{sec:conclusions}.

\section{The forward-forward algorithm} \label{sec:forward-forward}
The forward-forward algorithm~\cite{hinton2022forward} replaces the need for backpropagation by training each layer on its own. 
This also means that each layer is trained using its own loss function. 
The heuristic proposed by Hinton compares the activity in the layer for \emph{positive} and \emph{negative} samples.
A goodness function~\cite{hinton2022forward} is defined as follows and describes the probability of any given sample to be classified as a \emph{positive} sample:

\begin{equation}
    p(positive) = \sigma \Bigg(\sum_j y_j^2 - \theta\Bigg).
\end{equation}

In this case, $y_j$ is the activity of the neuron, $\sigma$ the logistic function, and $\theta$ some threshold.

For training the loss function of a layer $i$, with layer-activations for positive and negative samples being $y_p$ and $y_n$ respectively, might look like this:

\begin{equation}
    \mathcal{L}_i = \log(1 + \exp(-\norm{y_{p}} + \theta)) + \log(1 + \exp(\norm{y_{n}} - \theta))
\end{equation}

Where $\norm{y}$ refers to the $l_2$ norm of the vector $y$.

Intuitively, positive samples should be \emph{correct/good} images, while negative samples should be \emph{wrong/bad}. 
Where positive samples should induce high activity (\ie goodness) in the neurons, negative samples should lead to low activity of the neurons. (Although this condition can also be inverted). This means, the network is trained to identify samples that would fit into the dataset, while rejecting samples that do not.

\subsection{Supervised learning}\label{sec:ff-deep}

\begin{figure}
    \centering
    \includegraphics[width=0.3\linewidth]{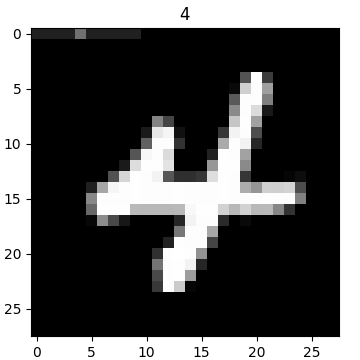}
    \caption{MNIST example of a sample with embedded label}
    \label{fig:sample_mnist_four}
\end{figure}

To facilitate supervised learning, we embed labels into the input image (see Figure~\ref{fig:sample_mnist_four}). 
To generate positive samples we replace the first $n$ pixels of the image with the one-hot encoded label of the sample (with $n$ being the number of classes). For negative samples, we replace the same pixels but embed a random, incorrect class instead.
In this context, the network is tasked to directly identify whether the embedded label fits the sample.

For classifying samples one can either use a softmax layer trained on the outputs of the network (\emph{fast} method) or compare the \emph{goodness} score of each label (embedded into the current sample) and pick the highest one (\emph{slow} method). 


When training multiple layers using this goodness-based loss, it is necessary to normalize the activations of each layer, before passing them to the next one. This ensures that the following layer cannot simply \emph{copy} the activations of the previous layer, to determine the goodness of the sample, but instead has to interpret the direction of the input vector.

\subsection{Other variants}
Hinton also proposed an unsupervised variant and a recurrent variant of the forward-forward algorithm.
For the unsupervised variant, negative samples are generated by combining multiple samples using a (modified) random mask. The positive samples are the unmodified samples of the dataset. A network trained using this method achieves similar performance to the supervised approach.

The recurrent variant is based on GLOM~\cite{hinton2022represent} and introduced to recreate the ability of backpropagation to pass information from later layers to earlier ones. In doing so, the recurrent variant circumvents one of the larger possible weaknesses of the forward-forward training algorithm. However, since it is a recurrent network it requires multiple passes, slowing down the training. 

For the purposes of this paper we will concentrate on the basic supervised training procedure, which can be easily used to pretrain on a given SSL task. However, we compare against the unsupervised approach in Section~\ref{sec:unsupervised_compare}.

\section{Self-Supervised Representation Learning}
\label{sec:SSL}

Self-supervised representation learning (SSL) refers to training procedures that pretrain a network on some task to learn representations that can be used to easily solve a variety of transfer tasks. 
There have been a number of works in this area over the last few years. 
Siamese networks in particular have been quite successful in producing good results. 

Siamese networks are generally trained to identify two augmented instances of the same images, while still being able to discriminate between different images. This is either achieved by a contrastive training procedure (\eg SimCLR~\cite{simclr}, MoCo~\cite{mocov1}) or introduction of asymmetry in the architecture (\eg SimSiam~\cite{simsiam}, BYOL~\cite{byol}).

In contrast to this are methods that operate on a single view of an image. 
These can be separated into generative approaches such as Inpainting or Colorization~\cite{zhang2016colorful} and into discriminative approaches which predict augmentations of the image such as Rotation~\cite{gidaris2018unsupervised}, Jigsaw~\cite{chen2021jigsaw,noroozi2016unsupervised} or Context Prediction~\cite{doersch2015unsupervised}.
While Siamese methods generally perform quite well, their approach of comparing two views cannot be easily mapped to the forward-forward algorithm, since early layers will be heavily affected by the augmentations employed in these methods. 

Similarly, we will not cover any generative approaches, as this would require an extension of the forward-forward algorithm with generative capabilities. How generative versions of forward-forward such as \cite{ororbia2023learning,ororbia2023predictive} could be applied to SSL-tasks like Inpainting and Colorization~\cite{zhang2016colorful} is left for future work to investigate.

\section{Experimental setup}
\label{sec:experimental_setup}
\subsection{Neural Network}
We generally follow the neural network setup as outlined by Hinton~\cite{hinton2022forward}. 
However, we stick to the most basic setup for all datasets, which consists of a 4-layer (ReLU-)MLP with 2000 neurons each.
Before each of the hidden layers, we include a normalization layer as it is required for deep learning of the forward-forward approach (see Section~\ref{sec:ff-deep}).
The SSL-task provides the labels that get embedded into the input image as described in Section~\ref{sec:forward-forward}.

For evaluation of transfer performance, we train a linear classifier on top of the trained network. 
The network is pretrained for 60 epochs, the linear classifier is then trained for 100 epochs on the frozen backend.
All networks are trained using the Adam~\cite{adam} optimizer with a fixed learning rate of $0.0001$.

We provide results using the forward-forward training algorithm, as well as comparisons to networks trained using backpropagation and a cross-entropy loss. Results trained with backpropagation and cross-entropy loss use the same settings and network architecture as the other experiments, with the addition of a final (10-neuron) layer which outputs the logits used for classification.

\subsection{Datasets}
We evaluate the performance of the various configurations on the following four datasets: MNIST~\cite{mnist}, Fashion-MNIST (F-MNIST)~\cite{fashion_mnist}, Street View House Numbers (SVHN)~\cite{svhn} and CIFAR-10~\cite{cifar}.
MNIST and CIFAR-10 are the datasets Hinton used for his experiments, while the F-MNIST and SVHN datasets are our addition.

Both MNIST and F-MNIST are very simple grayscale datasets with $28\times 28$ pixels images. 
We use the cropped version of the SVHN dataset, which means that the SVHN, as well as the CIFAR-10 datasets, both consist of $32\times 32$ color images.
All of these datasets consist of 10 different classes.

Note that an MLP trained on CIFAR-10 will typically overfit. However, since we are evaluating SSL-transfer performance we will include CIFAR-10 results as well. While overfitting on a pretext task is likely to lead to worse transfer performance, we still hope to gain some insights into how the forward-forward algorithm is affected by overfitting.

\subsection{SSL-Tasks}

As mentioned before, not every representation learning technique can be easily applied to the forward-forward algorithm. 
For this paper, we selected three SSL-tasks that can be formulated as a supervised classification task. 

Some of these tasks rely on assumptions that might not hold for every sample in every dataset, in which case a lower accuracy is to be expected.
For example, rotating the MNIST-digit \emph{zero} might not be very helpful.
In addition to the following SSL-tasks, we also train networks on the supervised classification task, to provide a baseline. 

The labels of the SSL-task are embedded into the input image, as described in Section~\ref{sec:forward-forward}. For multi-channel images (SVHN and CIFAR-10) we embed the labels into the first color-channel. 
\vspace{-0.2cm}
\paragraph{Rotation}
Learning to predict the rotation of an image as a pretraining task in the context of SSL has been explored by Gidaris~\etal~\cite{gidaris2018unsupervised}. Images are rotated in steps of 90° and the network is trained to predict this rotation. This translates to a SSL-task with 4 labels.
\vspace{-0.25cm}
\paragraph{Flipping}
Closely related to rotation, a network can be trained to predict horizontal (h) and/or vertical (v) flips of an image. The resulting task has 2 or 4 labels.
\vspace{-0.25cm}
\paragraph{Jigsaw}
Solving jigsaw puzzles is another quite popular approach to SSL~\cite{chen2021jigsaw,doersch2015unsupervised,noroozi2016unsupervised}. 
Applying this task to our framework of a single MLP without any weight-sharing, we split the image into 4 patches. The network is tasked to predict the permutation of the input image, with each possible permutation being encoded as a one-hot label.
When images are split into more patches (\ie $9! = 362,880$), choosing a different method to embed the labels than one-hot encoding would be necessary.
Alternatively one could also follow the work of Noroozi and Favaro~\cite{noroozi2016unsupervised} and only use a subset of all possible permutations.

\section{Experimental Results}
\label{sec:results}

We provide the accuracy achieved in the various SSL-tasks, as well as accuracy values when using the pretrained network on the final classification task. Note that we report slightly lower numbers than were observed by Hinton (for FF, as well as BP). We assume these are due to minor differences in implementation and tuning. 

\subsection{Pretraining on the SSL-task}
In Table~\ref{tab:ssl-acc} we present the accuracy of networks trained using the forward-forward approach and networks trained using backpropagation (BP) on several SSL-tasks.

For backpropagation accuracy is calculated as usual. For the forward-forward approach we compare the goodness scores of every possible label for each sample averaged over the last 3 layers. 
This \emph{slow} method to measure accuracy often gives slightly better results than the linear classifier. (See Section~\ref{sec:forward-forward})

In the supervised classification task the forward-forward algorithm generally gets quite close in performance to the backpropagation baselines.
(In our case $98.03\%$ vs $98.77\%$, as seen in Table~\ref{tab:classification})
This is also the case for most SSL-tasks, although for some of the more difficult tasks (\ie jigsaw) and datasets (\ie SVHN and CIFAR-10) the gap increases noticeably.

\subsection{Transfer performance}
We evaluate the transfer performance of every network by training a linear classifier on top of the last 3 layers of the pretrained network.

In Table~\ref{tab:classification} we first report the baseline supervised classification accuracy for both the forward-forward approach and backpropagation (BP). 
Note, that the network trained using forward-forward includes accuracy values obtained using the \emph{fast} method of querying a linear classifier, as well as the \emph{slow} method of computing the goodness for each label (\textrightarrow no LC).

Looking at the transfer performance of the various SSL tasks (see Table~\ref{tab:classification}), the forward-forward approach is clearly outperformed by backpropagation. Networks trained in a supervised manner using the forward-forward approach seem to specialize significantly more on the task they are trained on, discarding any information not necessary to the current objective. On the other hand, networks trained using backpropagation seem to retain more information that can be used to transfer from the original SSL-objective to the classification task.

Interestingly, with the exception of CIFAR-10, the forward-forward algorithm achieves its best transfer performance on the easiest SSL-task: horizontal flipping. 
This effect could be a consequence of how easy this task is. If the forward-forward algorithm is able to achieve optimal performance very quickly, it will not continue to learn and therefore not degrade features only needed for the transfer task to the same degree as we have seen for the other SSL-tasks.
For flipping on CIFAR-10, the performance difference is likely due to overfitting, as we will discuss in Section~\ref{sec:overfitting}.

Also notable is the significantly worse transfer performance for the jigsaw task on MNIST. This behavior might be due to the structure of the MNIST dataset. Since every image has a border of black pixels, it is possible to solve the jigsaw puzzle by comparing these pixels specifically. However, since these pixels are pretty much useless for classification, a network that ignores everything else might show this sort of behavior. 

\begin{table}[t]
    \setlength{\tabcolsep}{4pt}
    \begin{center}
    \begin{tabular}{lcccc}
    \toprule
        Method              & MNIST & F-MNIST       & SVHN  & CIFAR-10 \\
    \midrule
        Rotation            & 98.64 & 96.92         & 68.16 & 60.21 \\
        \textrightarrow BP  & 99.56 & 98.48         & 79.88 & 66.42 \\
        Flip h/v            & 97.14 & 82.84         & 50.37 & 37.63 \\
        \textrightarrow BP  & 99.05 & 88.21         & 67.73 & 41.24 \\
        Flip h              & 99.06 & 86.83         & 77.94 & 52.26 \\
        \textrightarrow BP  & 99.39 & 89.19         & 84.90 & 51.74 \\
        Jigsaw              & 96.85 & 91.93$^{*}$       & 52.67 & 59.26 \\
        \textrightarrow BP  & 99.86 & 98.91         & 97.35 & 93.81 \\
    \bottomrule
    \end{tabular}
    \end{center}
    \caption{Test accuracy (\%) on SSL-tasks, comparing forward-forward with backpropagation. \scriptsize{$^{*}$ used a lower learning rate, to prevent a problem with exploding loss.}}
    \label{tab:ssl-acc}
    \vspace{-0.5cm}
\end{table}

\begin{table}[t]
    \setlength{\tabcolsep}{4pt}
    \begin{center}
    \begin{tabular}{lcccc}
    \toprule
        Method                      & MNIST & F-MNIST       & SVHN  & CIFAR-10 \\
    \midrule
        Supervised                  & 93.79 & 87.31         & 68.77 & 47.60 \\
        \textrightarrow no LC       & 98.03 & 86.81         & 72.78 & 46.66 \\
        \textrightarrow BP          & 98.77 & 90.52         & 85.08 & 57.72 \\
        Rotation                    & 76.33 & 78.98         & 49.18 & 43.86 \\
        \textrightarrow BP          & 96.81 & 88.59         & 78.81 & 50.20 \\
        Flip h/v                    & 80.88 & 79.34         & 54.22 & 39.05 \\
        \textrightarrow BP          & 97.20 & 86.88         & 81.50 & 48.78 \\
        Flip h                      & 86.01 & 79.28         & 55.69 & 30.15 \\
        \textrightarrow BP          & 96.02 & 86.88         & 78.47 & 48.42 \\
        Jigsaw                      & 51.18 & 75.76$^{*}$       & 33.13 & 40.15 \\
        \textrightarrow BP          & 96.40 & 87.12         & 72.32 & 51.30 \\
    \bottomrule
    \end{tabular}
    \end{center}
    \caption{Test accuracy (\%) of classification transfer performance using a linear classifier, comparing forward-forward with backpropagation. \scriptsize{$^{*}$ used a lower learning rate, to prevent a problem with exploding loss.}}
    \label{tab:classification}
    \vspace{-0.5cm}
\end{table}

\subsection{Comparison of different losses for the backpropagation baseline}
The backpropagation baselines reported in Tables~\ref{tab:ssl-acc} and~\ref{tab:classification} were achieved using the Cross-Entropy loss. We also report results for networks trained with backpropagation using the same goodness-based loss that is used for the forward-forward algorithm.

In Table~\ref{tab:bp-loss} we compare the results of using different loss-functions for backpropagation for the supervised classification task, as well as the rotation prediction task. 
While Cross-Entropy (CE) loss performs best, networks trained using the Goodness-based loss get very close in performance. 

It is also possible to apply the Goodness-loss to every layer, which is more in line with the way the forward-forward algorithm optimizes each individual layer. (Although for BP the gradients will flow throught the network as usual).
Backpropagation achieves good results here as well (Goodness$_{all}$).
However, this approach has a significantly reduced transfer performance, which might explain part of the weak transfer performance of the forward-forward approach. Optimizing each layer with its own specific loss function, seems to generally result in the network discarding information not necessary to the current task.

This suggests that networks that were (pre-)trained using the supervised forward-forward algorithm require SSL-tasks that cannot be solved without retaining a lot of information that will be useful to the downstream task (\eg classification). 
As such the unsupervised training procedure proposed by Hinton achieves significantly better performance here. 

\begin{table}[t]
    \begin{center}

    \setlength{\tabcolsep}{4pt}
    \begin{tabular}{lcccc}
        \toprule
        Loss            & MNIST & F-MNIST       & SVHN  & CIFAR-10 \\
        \midrule
        \multicolumn{5}{c}{Supervised Classification} \\
        \midrule
        CE              & 98.76 & 89.86         & 83.98 & 56.42 \\ 
        Goodness   & 98.36 & 89.86         & 84.38 & 56.08 \\ 
        Goodness$_{all}$  & 98.32 & 89.93         & 82.48 & 55.79 \\ 
        \midrule
        \multicolumn{5}{c}{Rotation SSL Task} \\
        \midrule
        CE              & 99.56 & 98.48         & 79.89 & 66.42 \\ 
        Goodness   & 99.67 & 98.30         & 78.25 & 66.51 \\ 
        Goodness$_{all}$ & 99.35 & 98.23         & 76.93 & 65.13 \\ 
        \midrule
        \multicolumn{5}{c}{Rotation Transfer Performance} \\
        \midrule
        CE              & 96.82 & 88.59         & 78.81 & 50.21 \\ 
        Goodness        & 96.68 & 88.04         & 77.99 & 52.21 \\ 
        Goodness$_{all}$  & 68.74 & 78.69         & 67.11 & 48.15 \\ 
        \bottomrule
    \end{tabular}
    \end{center}
    \caption{Backpropagation results of training a network for classification and rotation. Comparing different possible loss functions. Transfer performance is evaluated by training a linear classifier on top of the pretrained network.}
    \label{tab:bp-loss}
    \vspace{-0.5cm}
\end{table}

\subsection{Overfitting on CIFAR-10}\label{sec:overfitting}
As expected, the selected network architecture is not very well suited for training on CIFAR-10. 
Especially networks trained directly for supervised classification overfit badly. 
Training on the various SSL tasks will also lead to overfitting, with some tasks (\eg jigsaw) overfitting less than others. 
Horizontal flipping shows the worst signs of overfitting on CIFAR-10, although in this specific case, the forward-forward algorithm drops less in accuracy moving from the train to the test split. This might explain why forward-forward is able to beat the transfer performance of backpropagation in this setting.
The same behavior (to a lesser degree) can also be observed on the SVHN dataset.

\section{Discussion}
\label{sec:discussion}

In order to gain a better understanding of how the forward-forward approach works for SSL-tasks, we generate visualizations of the embedding space using t-SNE~\cite{tsne}.
Following our existing experimental setup, we interpret the embedding of the outputs of the last 3 layers of the network.  (\ie every layer, but the first).

\begin{figure}[t]
    \centering
    \begin{subfigure}{.49\linewidth}
        \centering
        \includegraphics[width=\linewidth]{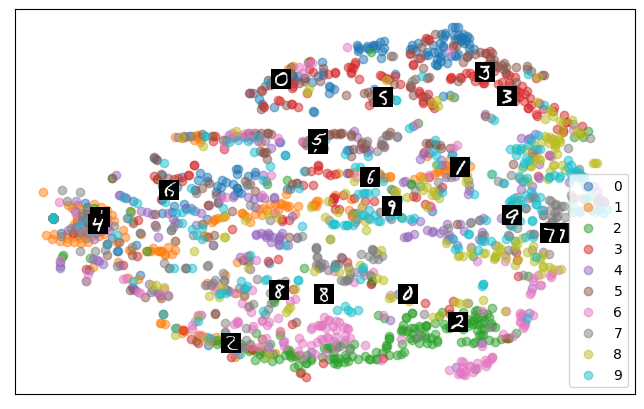}
        \caption{forward-forward}
        \label{fig:rotation_transfer_embedding_ff}
    \end{subfigure}
    \hfill    
    \begin{subfigure}{.49\linewidth}
        \centering
        \includegraphics[width=\linewidth]{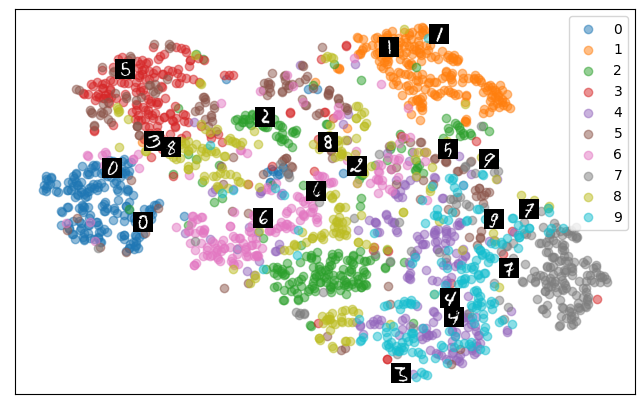}
        \caption{BP + Cross-Entropy loss}
        \label{fig:rotation_transfer_embedding_bp_ce}
    \end{subfigure}
    \begin{subfigure}{.49\linewidth}
        \centering
        \includegraphics[width=\linewidth]{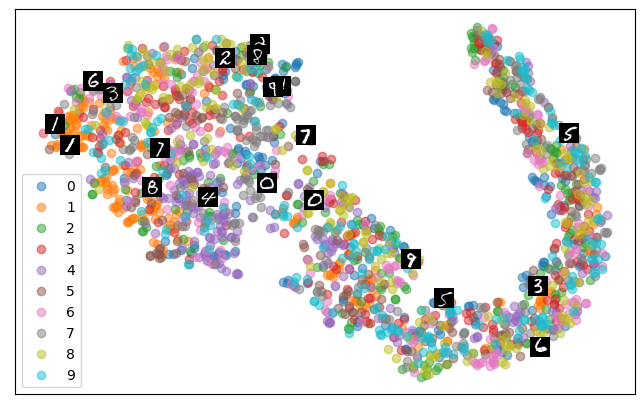}
        \caption{BP + per layer goodness loss}
        \label{fig:rotation_transfer_embedding_bp_g_all}
    \end{subfigure}
    \hfill
    \begin{subfigure}{.49\linewidth}\hfill
        \centering
        \includegraphics[width=\linewidth]{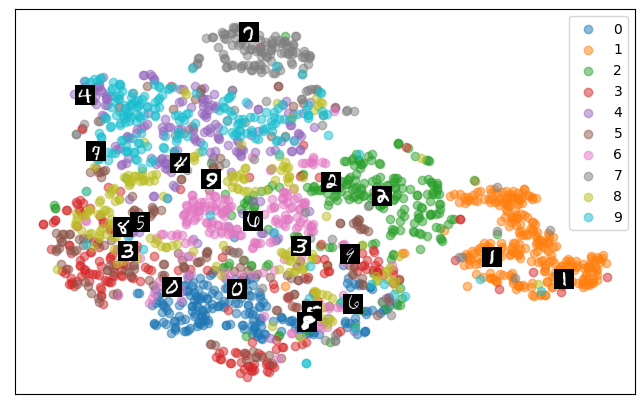}
        \caption{BP + goodness loss on last layer}
        \label{fig:rotation_transfer_embedding_bp_g_single}
    \end{subfigure}

    \caption{t-SNE-Visualization of the embedding of the MNIST digits by a network pretrained on the rotation SSL-task}
    \vspace{-0.5cm}

\end{figure}

In Figure~\ref{fig:rotation_transfer_embedding_ff} and Figure~\ref{fig:rotation_transfer_embedding_bp_ce} we take a look at the embedding-space of the MNIST digits for a network pretrained on the rotation SSL-task. 
As somewhat expected due to the observed transfer performance, the network trained using backpropagation has an embedding space that clusters the different digits significantly better, than the network trained by forward-forward. 

As discussed before, part of this behaviour can be attributed to the fact that every layer has its own loss function. Comparing these results to a network trained using backpropagation and the same per-layer goodness loss in Figure~\ref{fig:rotation_transfer_embedding_bp_g_all}, the same observation can be made. 

This indicates a major shortcoming of the forward-forward approach when applied to known SSL-techniques. Learning each layer using its own loss function seems to reduce the ability of a network to retain some of the information that might be \emph{less-relevant} to the current SSL task.
This assumption holds, when comparing against Figure~\ref{fig:rotation_transfer_embedding_bp_g_single}, which is the resulting embedding of a network trained using backpropagation and a goodness-based loss applied only to the last layer.

\subsection{Structure of the learned representations}
\begin{figure}[t]
    \centering
    \begin{subfigure}{0.49\textwidth}
        \centering
        \includegraphics[width=\textwidth]{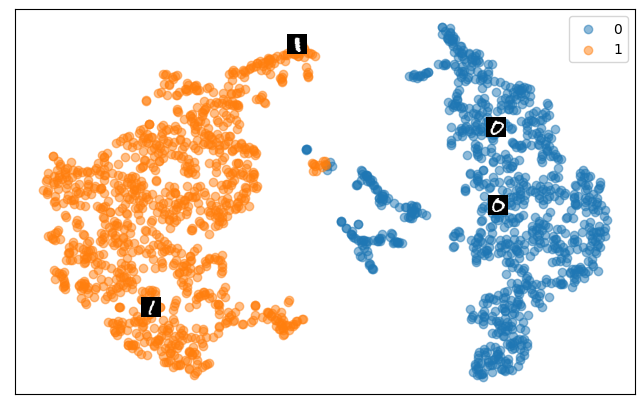}
        \caption{forward-foward}
        \label{fig:class_ff_mnist_2class}
    \end{subfigure}
    \hfill
    \begin{subfigure}{0.49\textwidth}
        \centering
        \includegraphics[width=\textwidth]{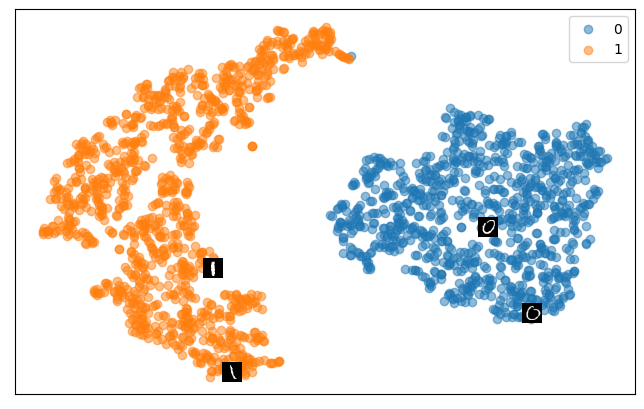}
        \caption{backpropagation}
        \label{fig:class_bp_mnist_2class}
    \end{subfigure}
    \caption{t-SNE-Visualization of the embedding generated by the last three layers for the MNIST digits zero and one by a network trained for supervised classification.}
    \vspace{-0.1cm}
\end{figure}

When training on a simplified classification task the different behavior between backpropagation and forward-forward become a bit more apparent. In this case the networks visualized in Figure~\ref{fig:class_ff_mnist_2class} and~\ref{fig:class_bp_mnist_2class} were trained on a reduced MNIST dataset, which was limited to the digits \emph{zero} and \emph{one}. 
While the network trained using backpropagation achieves two easily separable clusters of samples, the forward-forward approach features an area where no clear distinction is possible, even though most of the samples can be easily grouped into the two classes. 

\subsection{Representations of different layers}
In contrast to backpropagation, where the last layer is almost always the best suited to obtain a useful representation of the input, representations obtained from a forward-forward-trained network should in most cases include the outputs of intermediate layers as well~\cite{hinton2022forward}. Depending on the task, earlier layers can even be better suited for classification than later layers.

\begin{figure}[t]
    \centering
    \begin{subfigure}{0.49\textwidth}
        \centering
        \includegraphics[width=\textwidth]{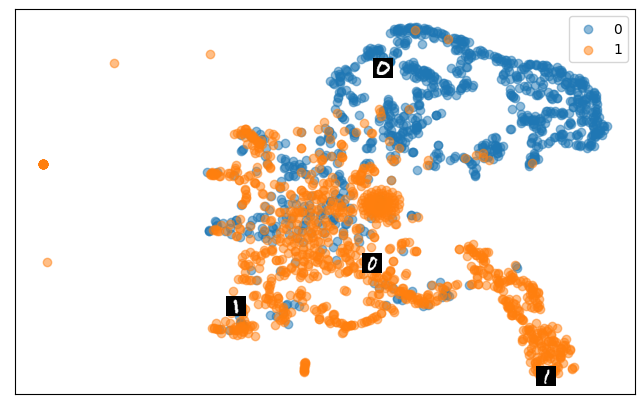}
        \caption{Using the last layer}
        \label{fig:rotation_transfer_mnist_2class_1_layer}
    \end{subfigure}
    \hfill
    \begin{subfigure}{0.49\textwidth}
        \centering
        \includegraphics[width=\textwidth]{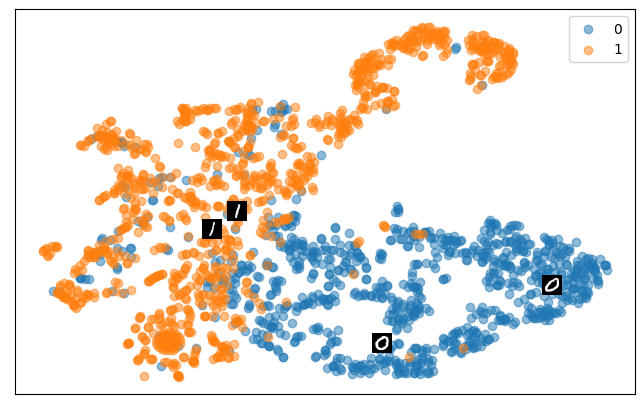}
        \caption{Using the last three layers}
        \label{fig:rotation_transfer_mnist_2class_3_layer}
    \end{subfigure}
    \caption{t-SNE-Visualization of the embedding generated for the MNIST digits zero and one by a network pretrained on the rotation SSL-task using FF training.}
    \label{fig:rotation_transfer_mnist_2class}
    \vspace{-0.5cm}
\end{figure}

For example, the network in Figure~\ref{fig:rotation_transfer_mnist_2class} was pretrained on the rotation SSL-task using a 2-class subset of MNIST (digits \emph{zero} and \emph{one}).
Where Figure~\ref{fig:rotation_transfer_mnist_2class_1_layer} shows the embedding of the last layer, Figure~\ref{fig:rotation_transfer_mnist_2class_3_layer} shows the embedding of the last three layers. 
While both embeddings tend to group the different digits, using the last three layers slightly improves the quality of this clustering. Similarly, including the last three layers (instead of only the last) as input for the linear classifier that is trained on the final classification task slightly improves accuracy.

Notably, this effect can be observed for networks pretrained using backpropagation as well, when using a Goodness loss.
For a deeper look into these effects in forward-forward learning, refer to the supplementary material.

\subsection{Comparison to unsupervised training}\label{sec:unsupervised_compare}
Our findings so far indicate that self-supervised representation learning using forward-forward has difficulty retaining information that is vital to the transfer task. 
The unsupervised training procedure, as described by Hinton~\cite{hinton2022forward}, sticks a bit closer to the overall idea of the forward-forward training.
Its training can be interpreted as a pretext task to discriminate between images from the dataset, and images outside of the dataset. 
Compared to the results we presented in this study, this unsupervised training procedure performs significantly better.

Interestingly the same procedure also achieves decent performance in predicting rotation, when trained on a dataset containing rotated images.
On MNIST it achieves $97.6\%$ test accuracy for rotation prediction (compared to $98.6\%$ when trained supervised). 
However, contrary to the network trained using SSL (which achieved $76.3\%$), the unsupervised model still achieves $97.9\%$ classification accuracy on MNIST.
These results support the intuition stated by Hinton~\cite{hinton2022forward}:
\begin{quote}
    ``Since the only difference between positive and negative data is the label, FF should ignore all features of the image that do not correlate with the label.''
\end{quote}
This also indicates that the way the supervised labels get embedded into the samples for the SSL-task might not be suited for pretraining and transfer learning.

\section{Conclusions}
\label{sec:conclusions}

In this paper, we investigated the performance of the newly proposed forward-forward training procedure in the context of self-supervised learning.
While the approach often performs competitively compared to the backpropagation baseline when trained for classification, the transfer performance from SSL-tasks in various tested configurations leaves much to be desired. 
Most of this behavior can be explained by the layer-wise loss in combination with the embedded labels, as a similar behavior can be observed when training a network using backpropagation with a similar setup. 

Generally, we observed that networks trained using supervised forward-forward on a SSL-task  tend to specialize strongly in that specific task, losing generalizability in the process. Considering that the unsupervised approach previously proposed by Hinton significantly outperforms the tested SSL methods, it seems to be the superior approach to learning without labels in a forward-forward setting. 
It remains to be seen whether ideas from known SSL methods can be applied successfully to the unsupervised case of the forward-forward algorithm, or whether a recurrent network is better suited for self-supervised learning.

\vspace{-0.3cm}
\paragraph{Future Work}
There are still quite a few approaches to self-supervised representation learning that do not map easily to the new forward-forward training procedure. Any techniques that rely on the network being able to generate images would require extensions to the basic formulation \cite{ororbia2023learning,ororbia2023predictive}. 
Similarly, the task of implementing Siamese network structures is left for future work, as they are quite reliant on strong image augmentation, that is not necessarily applicable to the smaller datasets. 

Since the forward-forward training procedure seems to often discard any information not relevant to the pretraining task, it becomes vital to choose a task that prevents this as much as possible or to change the forward-forward training procedure itself.

\section*{Acknowledgments}
This work was partly supported by The Alexander von Humboldt Foundation.
%
%
\bibliographystyle{splncs04}
\bibliography{egbib}

\clearpage


\appendix

\section{Embeddings of the different layers}\label{sec:embeddings_per_layer}

\begin{figure*}[!htb]
\begin{center}
    \begin{subfigure}{0.4\textwidth}
        \centering
        \includegraphics[width=\textwidth]{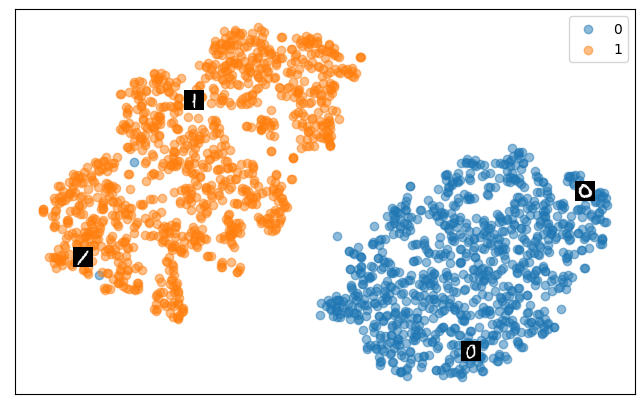}
        \caption{First layer}
        
    \end{subfigure}
    \begin{subfigure}{0.4\textwidth}
        \centering
        \includegraphics[width=\textwidth]{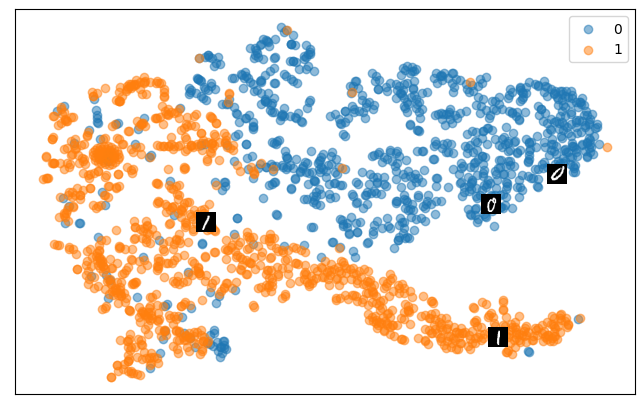}
        \caption{Second layer}
        
    \end{subfigure}
    \begin{subfigure}{0.4\textwidth}
        \centering
        \includegraphics[width=\textwidth]{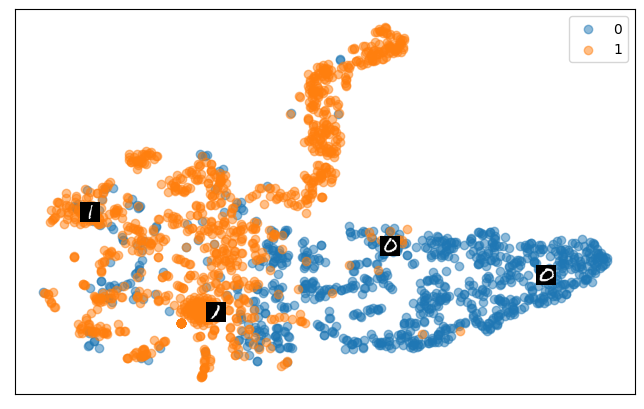}
        \caption{Third layer}
        
    \end{subfigure}
    \begin{subfigure}{0.4\textwidth}
        \centering
        \includegraphics[width=\textwidth]{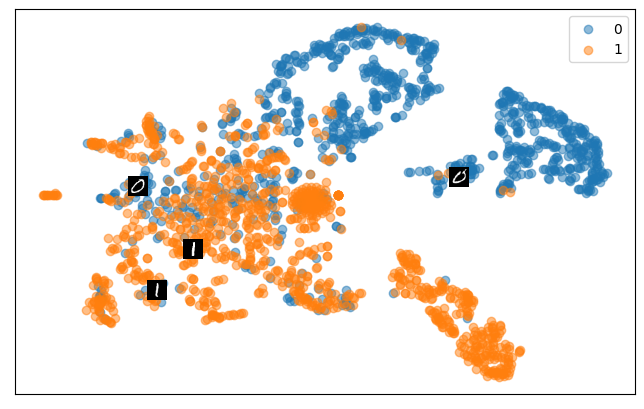}
        \caption{Fourth layer}
        \label{fig:tsne_2mnist_ff}
    \end{subfigure}
\end{center}
\vspace{-0.4cm}
\caption{t-SNE-embeddings for the digits zero and one as produced by a network pretrained (using forward-forward) on the rotation SSL-task, extracted at the given layer. Trained on a 2-class subset of the MNIST dataset}
\end{figure*}

Depending on the dataset and the task the network is trained for, the t-SNE embeddings \cite{tsne} can look quite different. 
Generally, the last layer of a network trained with forward-forward does the worst job clustering samples into their respective classes (see Figure \ref{fig:tsne_2mnist_ff}). This is especially apparent when pretraining on a SSL-task. However, looking at the (relatively easy) dataset of 2-class-MNIST, the earlier layers show better separability of the two classes.

\begin{figure*}[!htb]
\begin{center}
    \begin{subfigure}{0.4\textwidth}
        \centering
        \includegraphics[width=\textwidth]{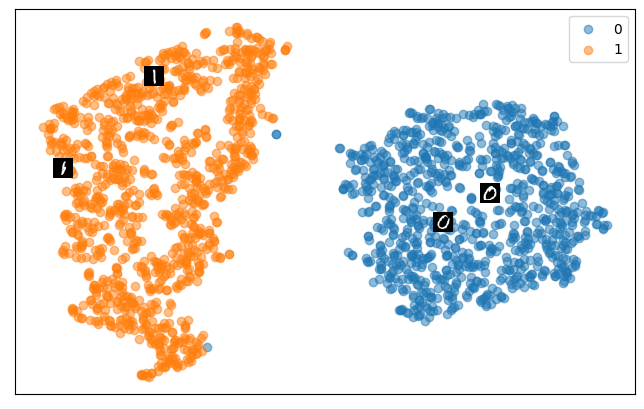}
        \caption{First layer}
        
    \end{subfigure}
    \begin{subfigure}{0.4\textwidth}
        \centering
        \includegraphics[width=\textwidth]{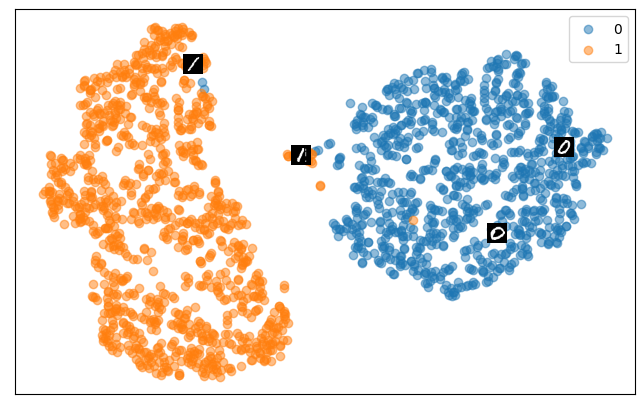}
        \caption{Second layer}
        
    \end{subfigure}
    \begin{subfigure}{0.4\textwidth}
        \centering
        \includegraphics[width=\textwidth]{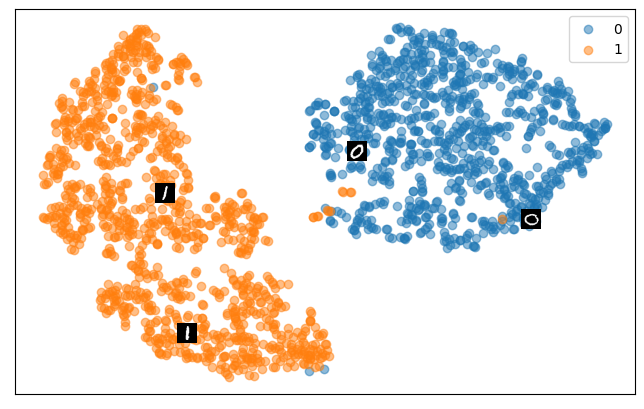}
        \caption{Third layer}
        
    \end{subfigure}
    \begin{subfigure}{0.4\textwidth}
        \centering
        \includegraphics[width=\textwidth]{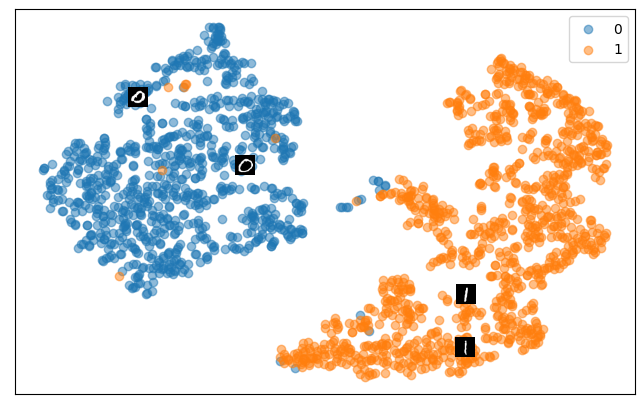}
        \caption{Fourth layer}
        
    \end{subfigure}
\end{center}
\vspace{-0.4cm}
\caption{t-SNE-embeddings for the digits zero and one as produced by a network pretrained (using BP with a cross-entropy loss) on the rotation SSL-task, extracted at the given layer. Trained on a 2-class subset of the MNIST dataset}\label{fig:tsne_2mnist_bp_ce}
\end{figure*}

\begin{figure*}[!htb]
\begin{center}
    \begin{subfigure}{0.4\textwidth}
        \centering
        \includegraphics[width=\textwidth]{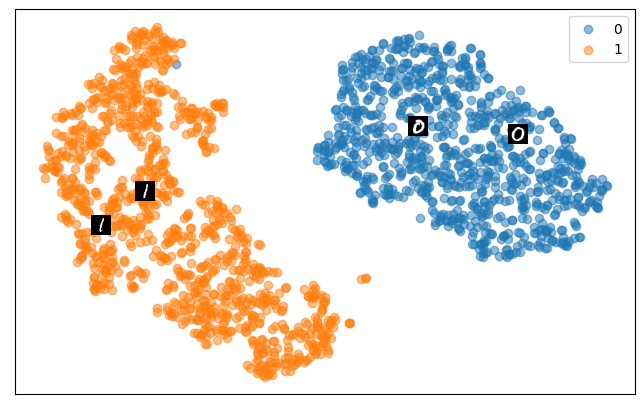}
        \caption{First layer}
        
    \end{subfigure}
    \begin{subfigure}{0.4\textwidth}
        \centering
        \includegraphics[width=\textwidth]{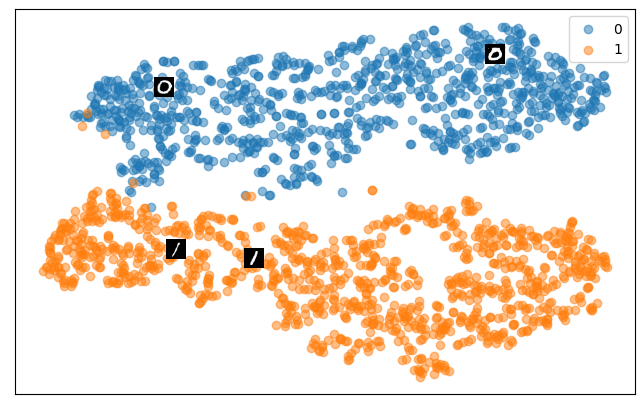}
        \caption{Second layer}
        
    \end{subfigure}
    \begin{subfigure}{0.4\textwidth}
        \centering
        \includegraphics[width=\textwidth]{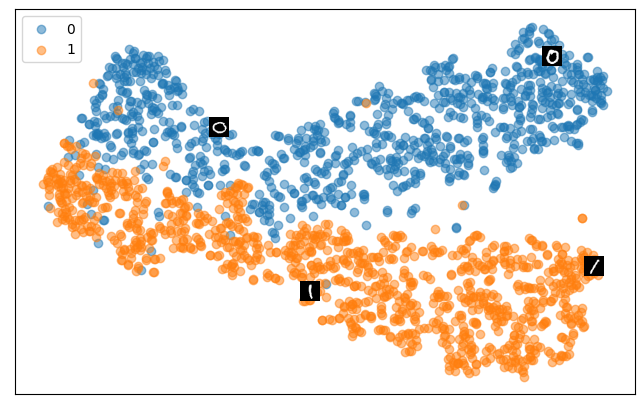}
        \caption{Third layer}
        
    \end{subfigure}
    \begin{subfigure}{0.4\textwidth}
        \centering
        \includegraphics[width=\textwidth]{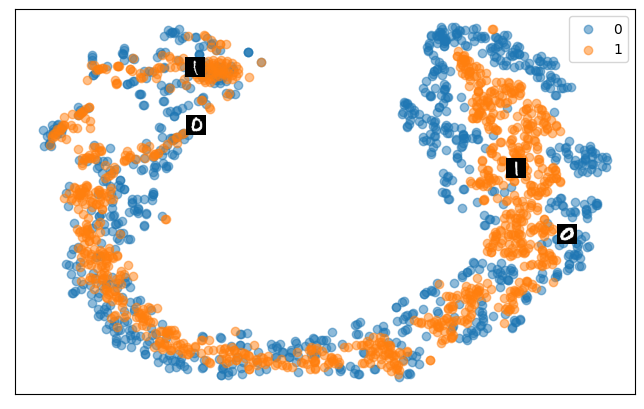}
        \caption{Fourth layer}
        
    \end{subfigure}
\end{center}
\vspace{-0.2cm}
\caption{t-SNE-embeddings for the digits zero and one of a network pretrained (using BP with a per-layer goodness loss) on the rotation SSL-task, extracted at the given layer. Trained on a 2-class subset of the MNIST dataset}\label{fig:tsne_2mnist_bp_goodness_all}
\end{figure*}

The same holds when training a network using backpropagation and the same goodness-loss for each layer (while still allowing gradients to flow from the last layer to the first nonetheless), as seen in Figure \ref{fig:tsne_2mnist_bp_goodness_all}. However, when looking at the embeddings of a network pretrained using backpropagation with a cross-entropy loss, as seen in Figure \ref{fig:tsne_2mnist_bp_ce}, the 2 classes are easily separable for all layers.

\begin{figure*}[!htb]
\begin{center}
    \begin{subfigure}{0.4\textwidth}
        \centering
        \includegraphics[width=\textwidth]{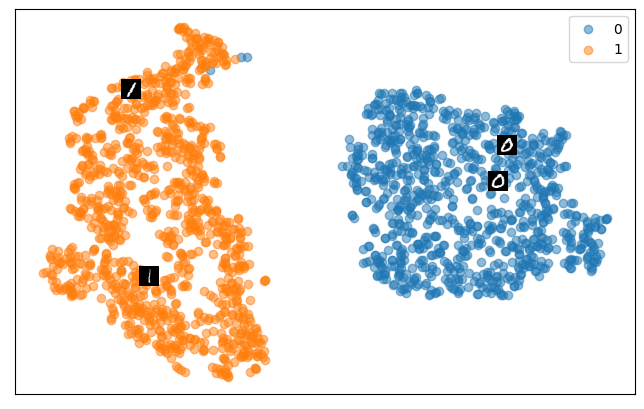}
        \caption{First layer}
        
    \end{subfigure}
    \begin{subfigure}{0.4\textwidth}
        \centering
        \includegraphics[width=\textwidth]{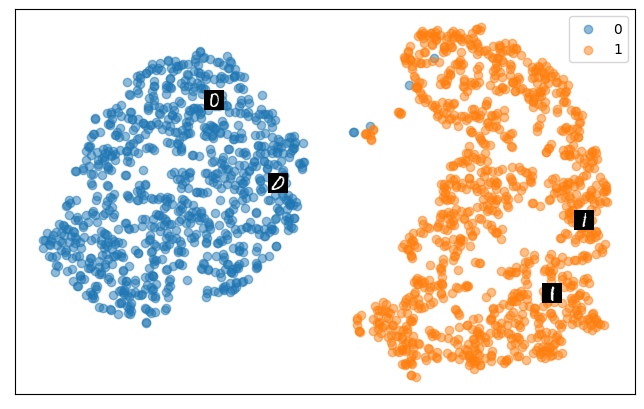}
        \caption{Second layer}
        
    \end{subfigure}
    \begin{subfigure}{0.4\textwidth}
        \centering
        \includegraphics[width=\textwidth]{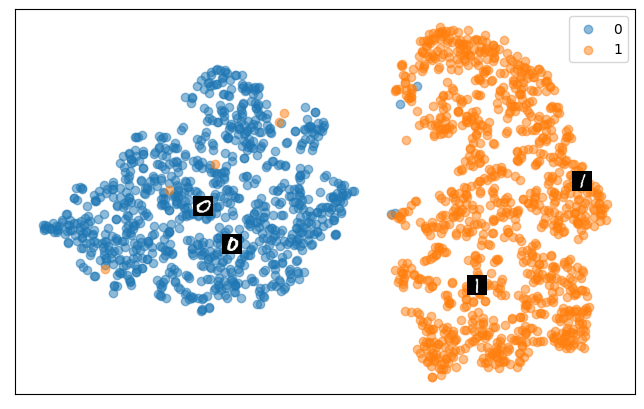}
        \caption{Third layer}
        
    \end{subfigure}
    \begin{subfigure}{0.4\textwidth}
        \centering
        \includegraphics[width=\textwidth]{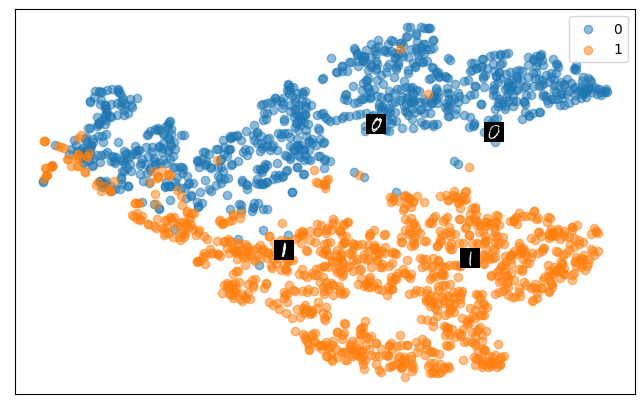}
        \caption{Fourth layer}
        
    \end{subfigure}
\end{center}
\vspace{-0.2cm}
\caption{t-SNE-embeddings for the digits zero and one as produced by a network pretrained (using BP with goodness loss for the last layer) on the rotation SSL-task, extracted at the given layer. Trained on a 2-class subset of the MNIST dataset}\label{fig:tsne_2mnist_bp_goodness}
\end{figure*}

Using goodness loss with backpropagation on only the last layer gives decent results as well (see Figure \ref{fig:tsne_2mnist_bp_goodness}), although the embedding generated when looking only at the last layer is slightly worse when compared to the cross-entropy case.

This might indicate that at least some of this \emph{specialization} behavior observed during our experiments can be explained by the selection of the loss function. 

Interestingly, while performance on the SSL-task usually is worst for the first layers (which is the reason we follow Hinton in using the output of the last 3 layers to measure accuracy), we observed a different behavior for some configurations when evaluating transfer performance.

For example, a forward-forward pretrained network on MNIST showed increased transfer performance when only sampling the first layer. Considering this is the layer that also performed worst on the SSL-task it was trained on, we assume that this layer \emph{discards} the least amount of information (that is not required to solve the SSL-task). 

\section{Supervised accuracy of different layers}
For networks trained using forward-forward we can measure the accuracy of the current supervised SSL-task by comparing the neuron activity.
While generally the best accuracy can be achieved by evaluating the goodness of all layers except the first, there can be subtle differences depending on the difficulty of the task and dataset.

In Table \ref{tab:layer_accuracies} we show compare accuracy for the various SSL-tasks when evaluating using the \emph{slow} method with only a single layer. Usually, the second layer of the network seems to perform the best. Only for the easiest dataset (MNIST), the first layer achieves better test accuracy for the SSL-tasks. This mostly coincides with the observations made by Hinton \cite{hinton2022forward} when evaluating the supervised forward-forward approach and is the reason we generally use the last three layers to calculate accuracy.

\begin{table}[]
    \centering
    \setlength{\tabcolsep}{2pt}
    \begin{tabular}{l|cccc|cccc}
    \toprule
    Task & 1\textsuperscript{st} layer & 2\textsuperscript{nd} layer & 3\textsuperscript{rd} layer & 4\textsuperscript{th} layer & 1\textsuperscript{st} layer & 2\textsuperscript{nd} layer & 3\textsuperscript{rd} layer & 4\textsuperscript{th} layer \\
    \midrule
    \multicolumn{5}{c}{MNIST} & \multicolumn{4}{c}{Fashion-MNIST} \\
    \midrule
    Classify  & 97.47 & 98.38 & 97.98 & 97.57 & 85.78 & 87.74 & 85.75 & 84.66\\
    Rotation        & 99.08 & 98.97 & 98.32 & 97.88 & 96.15 & 97.47 & 96.44 & 95.63 \\
    Flip h/v        & 97.70  & 97.41 & 96.85 & 96.05 & 81.37 & 83.77 & 81.99 & 80.34 \\
    Flip h          & 99.51 & 99.13 & 99.02 & 98.86 & 86.47 & 87.22 & 86.74 & 86.03 \\
    Jigsaw          & 98.37 & 98.23 & 91.91 & 87.86 & 86.92 & 87.99 & 86.42 & 85.67 \\
    \midrule
    \multicolumn{5}{c}{SVHN} & \multicolumn{4}{c}{CIFAR-10} \\
    \midrule
    Classify  & 85.73 & 95.78 & 94.96 & 93.75 & 42.91 & 48.36 & 43.29 & 38.68 \\
    Rotation        & 59.78 & 68.52 & 67.21 & 64.52 & 54.25 & 60.57 & 58.73 & 57.54 \\
    Flip h/v        & 42.06 & 50.96 & 49.89 & 48.28 & 42.06 & 50.96 & 49.89 & 48.28 \\
    Flip h          & 74.39 & 77.98 & 77.89 & 77.07 & 51.89 & 52.61 & 51.82 & 51.49 \\
    Jigsaw          & 29.82 & 72.34 & 64.16 & 52.67 & 35.39 & 64.74 & 54.35 & 40.34 \\
    \bottomrule
    \end{tabular}
    \caption{Per layer accuracy (\%), when evaluating the current supervised task using the \emph{slow} method.}
    \label{tab:layer_accuracies}
    \vspace{-1cm}
\end{table}

\section{Neuron activity when embedding labels}\label{sec:neuron_activity}

Since forward-forward networks are trained to have high activity when the correct label is embedded, and low activity when the incorrect label is embedded, we investigate how the generated t-SNE embedding behaves when embedding different labels.

\begin{figure*}[!htb]
\begin{center}
    \begin{subfigure}{0.34\textwidth}
        \centering
        \includegraphics[width=\textwidth]{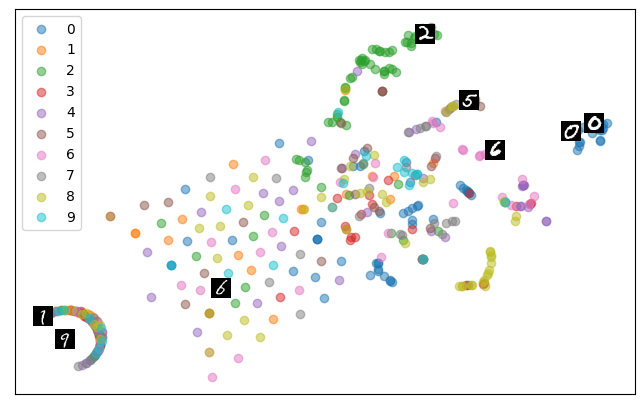}
        \caption{No label embedded.}
        \label{fig:neuron_activity_mnist_class:all}
    \end{subfigure}
    \hfill
    \begin{subfigure}{0.316\textwidth}
        \centering
        \includegraphics[width=\textwidth]{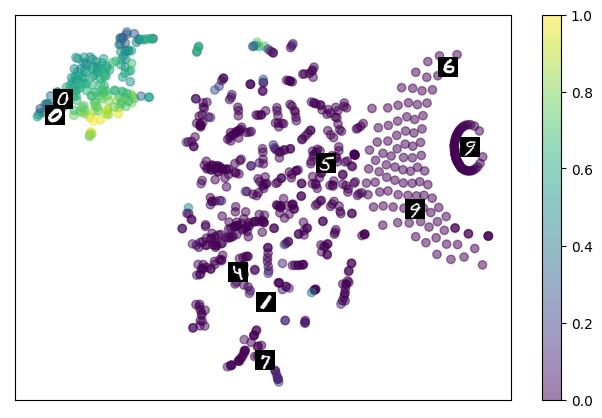}
        \caption{Label 0 embedded}
        \label{fig:neuron_activity_mnist_class:label_0}
    \end{subfigure}
    \hfill
    \begin{subfigure}{0.316\textwidth}
        \centering
        \includegraphics[width=\textwidth]{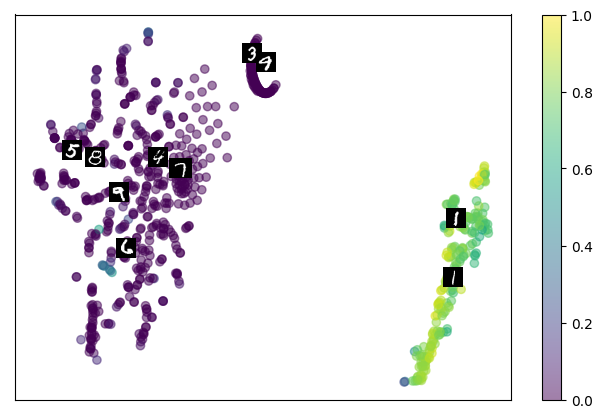}
        \caption{Label 1 embedded}
        \label{fig:neuron_activity_mnist_class:label_1}
    \end{subfigure}
\end{center}
\caption{t-SNE-embeddings for the MNIST digits as produced by a network trained on classification using forward-forward, extracted at the last layer. The color in the second and third plots refers to the (normalized) mean squared neuron activations.}
\label{fig:neuron_activity_mnist_class}
\vspace{-1cm}
\end{figure*}

As you can see in Figure \ref{fig:neuron_activity_mnist_class} the generated t-SNE embeddings depend heavily on the embedded label, when dealing with networks trained using supervised forward-forward. As expected, the samples of the correct label will get much higher activity in the layer, separating them into their own cluster of samples in this visualization (Figure \ref{fig:neuron_activity_mnist_class:label_0} and \ref{fig:neuron_activity_mnist_class:label_1}).

\begin{figure*}[!htb]
\begin{center}
    \begin{subfigure}{0.34\textwidth}
        \centering
        \includegraphics[width=\textwidth]{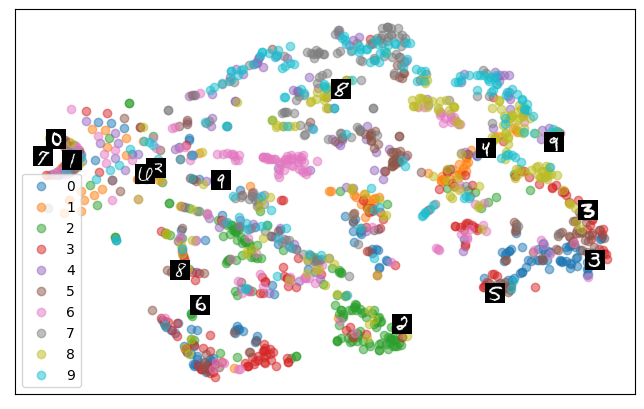}
        \caption{No label embedded.}
        \label{fig:neuron_activity_mnist_rotation:all}
    \end{subfigure}
    \hfill
    \begin{subfigure}{0.316\textwidth}
        \centering
        \includegraphics[width=\textwidth]{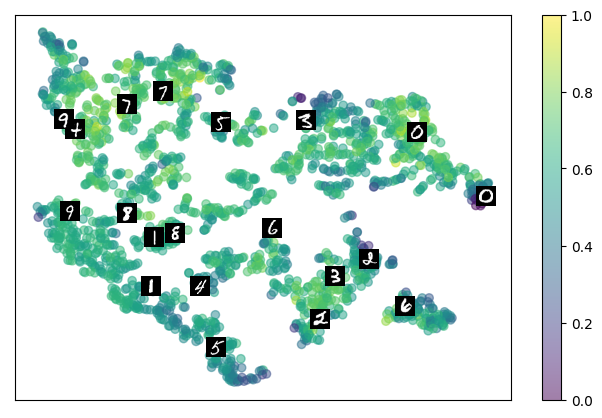}
        \caption{Label 0 embedded}
        \label{fig:neuron_activity_mnist_rotation:label_0}
    \end{subfigure}
    \hfill
    \begin{subfigure}{0.316\textwidth}
        \centering
        \includegraphics[width=\textwidth]{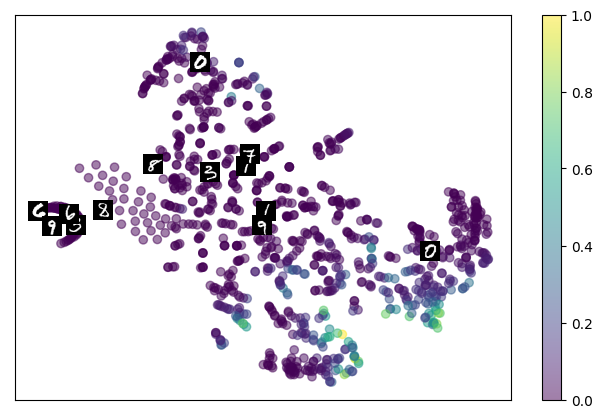}
        \caption{Label 1 embedded}
        \label{fig:neuron_activity_mnist_rotation:label_1}
    \end{subfigure}
\end{center}
\caption{t-SNE-embeddings for the MNIST digits as produced by a network pretrained on the rotation SSL-task using forward-forward, extracted at the last layer. The color in the second and third plots refers to the (normalized) mean squared neuron activations.}
\label{fig:neuron_activity_mnist_rotation}
\vspace{-1cm}
\end{figure*}

Note, that if no label is embedded, the samples are much harder to separate.
Obviously, this also means that when pretraining on a SSL-task and looking at the transfer (classification) performance this behavior can only be seen for the SSL-task (and not the transfer task).
As seen in Figure \ref{fig:neuron_activity_mnist_rotation} the neuron activity of a network pretrained using forward-forward on the rotation SSL-task behaves exactly as expected. In the context of the rotation SSL-task class 0 refers to not rotated, and class 1 refers to a rotation by 90 degrees. Therefore, the overall layer activity is much higher when embedding the 0 label (Since none of the samples are rotated).
Interestingly, when embedding the label 0, the t-SNE visualization seems to spread the samples more evenly, than when embedding the (in this context \emph{incorrect/negative}) label 1.

\begin{figure*}[!htb]
\begin{center}
    \begin{subfigure}{0.34\textwidth}
        \centering
        \includegraphics[width=\textwidth]{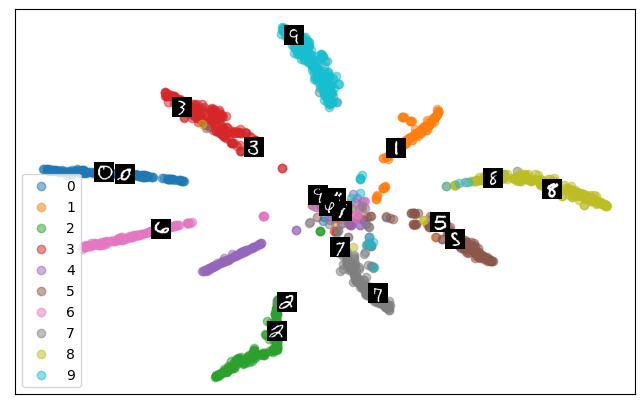}
        \caption{No label embedded.}
        \label{fig:neuron_activity_mnist_class_bp:all}
    \end{subfigure}
    \hfill
    \begin{subfigure}{0.316\textwidth}
        \centering
        \includegraphics[width=\textwidth]{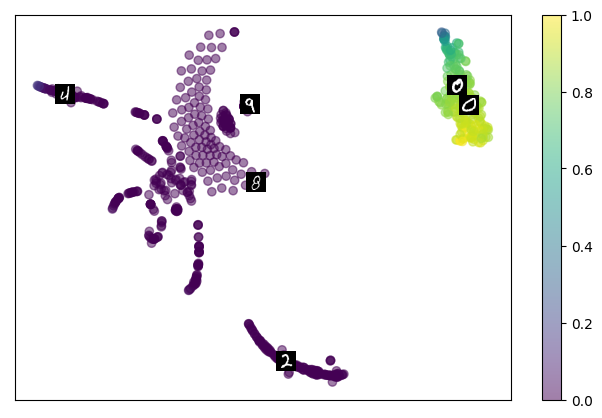}
        \caption{Label 0 embedded}
        \label{fig:neuron_activity_mnist_class_bp:label_0}
    \end{subfigure}
    \hfill
    \begin{subfigure}{0.316\textwidth}
        \centering
        \includegraphics[width=\textwidth]{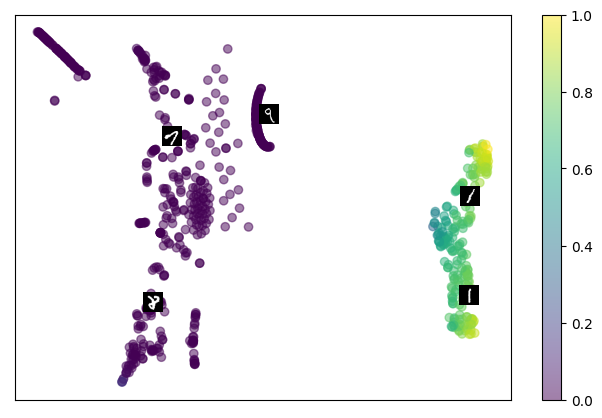}
        \caption{Label 1 embedded}
        \label{fig:neuron_activity_mnist_class_bp:label_1}
    \end{subfigure}
\end{center}
\caption{t-SNE-embeddings for the MNIST digits as produced by a network trained on classification using backpropagation and a goodness-less at the final layer, extracted at the last layer. The color in the second and third plots refers to the (normalized) mean squared neuron activations.}
\label{fig:neuron_activity_mnist_class_bp}
\vspace{-1cm}
\end{figure*}

Similarly, a network trained using backpropagation but using a goodness-loss at the last layer, behaves in a similar manner when looking at the last layer (See Figure \ref{fig:neuron_activity_mnist_class_bp}). 
Earlier layers do not show this behavior in this case, since the goodness-loss will affect the last layer the most. 
In contrast to forward-forward, the t-SNE embedding of a network trained with backpropagation for an image without an embedded label separates the classes much better (See Figure \ref{fig:neuron_activity_mnist_class_bp:all}).

\section{Embeddings of unsupervised training}
\begin{figure*}[!htb]
\begin{center}
    \begin{subfigure}{0.79\textwidth}
        \centering
        \includegraphics[width=\textwidth]{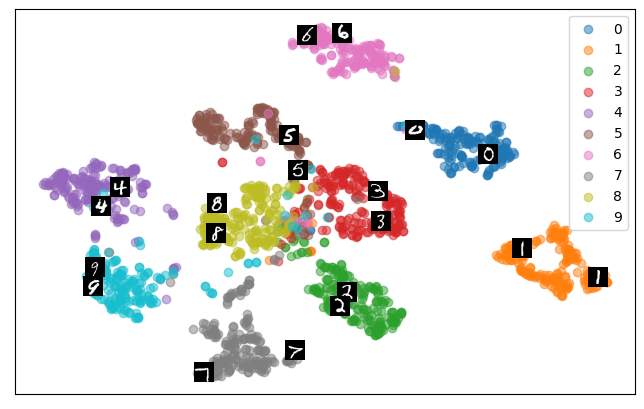}
        \caption{Trained on original MNIST}
        
    \end{subfigure}
    \begin{subfigure}{0.79\textwidth}
        \centering
        \includegraphics[width=\textwidth]{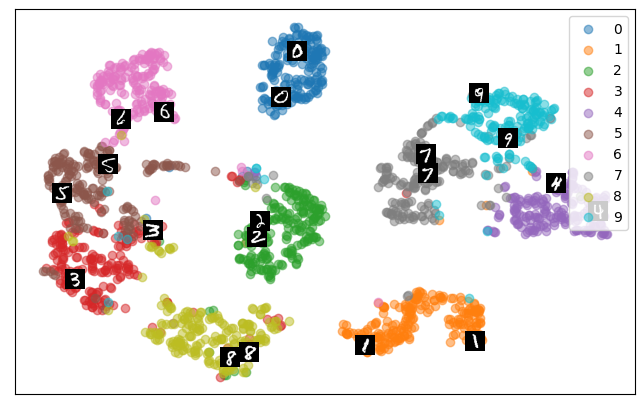}
        \caption{Trained with random rotation}
        
    \end{subfigure}
\end{center}
\vspace{-0.2cm}
\caption{t-SNE-embeddings for a network trained using unsupervised forward-forward on MNIST. Obtained from the last layer of the network.}\label{fig:tsne_unsupervised_mnist}
\end{figure*}

For completeness, we also provide t-SNE embeddings for networks trained using the forward-forward unsupervised learning method, as described by Hinton \cite{hinton2022forward}.
In Figure \ref{fig:tsne_unsupervised_mnist} we provide t-SNE embeddings for a network trained on the default MNIST, as well as a network trained on randomly rotated (0, 90, 180, 270 degrees) MNIST images.
The resulting t-SNE visualization generally clusters the different digits very well (similar to backpropagation). This also results (as noted in the paper) in decent classification accuracy.

\section{Behavior when training for longer}
To observe the effects of longer training we train networks for 200 epochs using backpropagation (with CE-loss) and forward-forward on the supervised classification task, as well as on the rotation SSL-task and compare classification accuracy for the MNIST digits.
Reported accuracy values are obtained by a linear classifier trained in parallel to the pretraining on the last 3 layers of the network.  

\begin{figure}
    \centering
    \begin{subfigure}{0.49\textwidth}
        \begin{tikzpicture}
        \begin{axis}[legend pos=south east, width=\textwidth,height=5cm, xlabel={Number of training epochs}, ylabel={Accuracy (\%)}, ymin=0, ytick={0,20,40,60,80,100}]
        \addplot table [x=Step, y=Value, col sep=comma, mark=none, y expr=\thisrow{Value} * 100] {forward-forward-ssl/data/classification_mnist_bp-ce.csv};
        \addplot table [x=Step, y=Value, col sep=comma, mark=none, y expr=\thisrow{Value} * 100] {forward-forward-ssl/data/classification_mnist_ff.csv};
        \legend{backpropagation, forward-forward}
        \end{axis}
        \end{tikzpicture}
        \caption{Supervised classification}
        \label{fig:cl_acc_plot}
    \end{subfigure}
    \hfill
    \begin{subfigure}{0.49\textwidth}
        \begin{tikzpicture}
        \begin{axis}[legend pos=south east, width=\textwidth,height=5cm, xlabel={Number of training epochs}, ylabel={Accuracy (\%)}, ymin=0, ytick={0,20,40,60,80,100}]
        \addplot table [x=Step, y=Value, col sep=comma, mark=none, y expr=\thisrow{Value} * 100] {forward-forward-ssl/data/cl_rotation_mnist_bp-ce.csv};
        \addplot table [x=Step, y=Value, col sep=comma, mark=none, y expr=\thisrow{Value} * 100] {forward-forward-ssl/data/cl_rotation_mnist_ff.csv};
        \legend{backpropagation, forward-forward}
        \end{axis}
        \end{tikzpicture}
        \caption{Transfer from rotation SSL-task}
        \label{fig:cl_rot_acc_plot}
    \end{subfigure}
    \caption{Classification accuracy vs. train epochs on MNIST classification}
    \vspace{-0.5cm}
\end{figure}
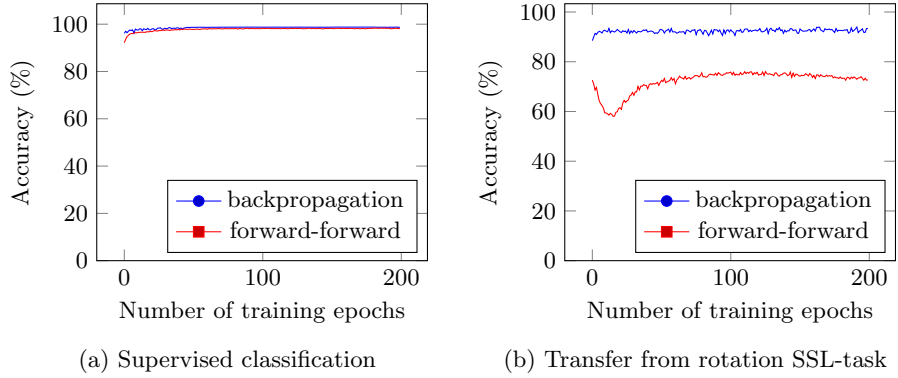

As already observed by Hinton, backpropagation usually converges slightly faster compared to forward-forward training. 
We observed the same effect in our experiments in Figure~\ref{fig:cl_acc_plot}. Even though the forward-forward approach does not achieve the same performance it comes quite close.

The transfer performance seen in Figure~\ref{fig:cl_rot_acc_plot} tells a similar story as previous observations. 
When training for longer the forward-forward approach even seems to begin dropping in transfer performance after some time. This is the case, even though performance on the SSL-task was still increasing throughout the whole experiment. 

This again indicates that the forward-forward approach will discard any information not correlated to the embedded label of the current SSL-task. So information that could be useful for downstream classification, but is not required for the SSL-task will get discarded, resulting in worse transfer performance.

\end{document}